\documentclass[conference]{IEEEtran}
\IEEEoverridecommandlockouts
\usepackage{cite}
\usepackage{amsmath,amssymb,amsfonts}
\usepackage{algorithmic}
\usepackage{graphicx}
\usepackage{textcomp}
\usepackage{xcolor}
\usepackage{hyperref}
\usepackage{cleveref}
\usepackage{booktabs}
\usepackage{acronym}
\usepackage{listings}
\usepackage{subcaption}
\usepackage{pifont}
\usepackage[inline]{enumitem}

\acrodef{fl}[FL]{federated learning}

\lstdefinelanguage{python}{
  keywords={import, from, as, class, def, return, lambda, yield, global, nonlocal, assert, 
  if, elif, else, try, except, finally, raise, while, for, in, break, continue, pass, 
  with, async, await, del, and, or, not, is, in, True, False, None},
  sensitive=true,
  columns=fullflexible,
  otherkeywords={...},
  morecomment=[l]{//},
  morecomment=[n]{/*}{*/},
  morestring=[b]",
  stringstyle=\ttfamily\color{red!50!brown},
  keywordstyle=\color{blue!80!black}\bfseries,
  keywordstyle=[2]\color{violet},
  commentstyle=\color{ddarkgreen}\itshape,
  showstringspaces=false,
  morestring=[b]',
  morestring=[b]""",
  basicstyle=\sffamily\lst@ifdisplaystyle\scriptsize\fi\ttfamily,
  emphstyle=\sffamily\bfseries\ttfamily
}
\definecolor{ddarkgreen}{rgb}{0,0.5,0}

\def\BibTeX{{\rm B\kern-.05em{\sc i\kern-.025em b}\kern-.08em
    T\kern-.1667em\lower.7ex\hbox{E}\kern-.125emX}}
\begin{document}

\title{ProFed: a Benchmark for Proximity-based non-IID Federated Learning}

\author{
    \IEEEauthorblockN{Davide Domini}
    \IEEEauthorblockA{
    \textit{
    University of Bologna}\\
    Cesena, Italy\\
    davide.domini@unibo.it
    }
    \and
    \IEEEauthorblockN{Gianluca Aguzzi} 
    \IEEEauthorblockA{
    \textit{
    University of Bologna}\\
    Cesena, Italy\\
    gianluca.aguzzi@unibo.it 
    }
    \and
    \IEEEauthorblockN{Mirko Viroli}
    \IEEEauthorblockA{
    \textit{
    University of Bologna}\\
    Cesena, Italy\\
    mirko.viroli@unibo.it 
    }
}


\maketitle

\begin{abstract}
In recent years, 
 \Ac{fl} has gained significant attention 
 within the machine learning community.
Although various \ac{fl} algorithms have been proposed in the literature, 
 their performance often degrades when data across clients is 
 non-independently and identically distributed (non-IID). 
This skewness in data distribution often emerges from geographic 
 patterns, with notable examples including regional linguistic variations
 in text data or localized traffic patterns in urban environments.
Such scenarios result in IID data within specific regions 
 but non-IID data across regions. 
However, existing \ac{fl} algorithms are typically evaluated by randomly splitting 
 non-IID data across devices, disregarding their spatial distribution.
    
To address this gap, we introduce ProFed, a benchmark that simulates data splits 
 with varying degrees of skewness across different regions. 
We incorporate several skewness methods from the literature and apply them to 
 well-known datasets, including MNIST, FashionMNIST, CIFAR-10, and CIFAR-100.
Our goal is to provide researchers with a standardized framework to evaluate FL algorithms 
 more effectively and consistently against established baselines.
\end{abstract}


\begin{IEEEkeywords}
Machine learning, Benchmarks, Performance evaluation
\end{IEEEkeywords}

\section{Introduction}\label{sec:intro}

\Acf{fl} has gained significant interests in the last years.
It has been introduced to address privacy problems 
 during learning from users data. 
In fact, this framework allows the training of a shared global model 
 without the need of collecting the data on a central server.

Different studies~\cite{DBLP:journals/fgcs/MaZLCQ22,DBLP:conf/aaai/HuangCZWLPZ21} 
 have shown that while \ac{fl} achieve good learning performance compared to classical learning approaches on homogeneously distributed data, 
 it drops performance when data are \emph{non-independently and identically distributed} (Non-IID).
For instance, in urban traffic prediction scenarios, data patterns exhibit strong 
 spatial correlations: traffic flows within specific city districts often share 
 common characteristics, while differing significantly from patterns observed in 
 other areas. This geographical dependency implies that models trained on data 
 from one district typically achieve higher accuracy when predicting traffic 
 patterns within the same area compared to predictions in different districts.

In the literature, various algorithms--like Scaffold~\cite{DBLP:conf/icml/KarimireddyKMRS20} 
 and FedProx~\cite{DBLP:conf/mlsys/LiSZSTS20}--have been proposed 
 to tackle data heterogeneity. 
These approaches typically assume that client data are distributed without considering specific 
 patterns or structures.
 However, in real world scenarios, particularly in the case of highly distributed 
  systems~\cite{DBLP:conf/acsos/Domini24} (e.g., in edge computing or spatial-aware scenarios), 
  it is common for data from geographically  close devices to be more similar to each other compared to data from devices 
  that are farther apart.
This phenomenon is driven by the fact that devices in the same region often experience 
 similar environments and make comparable observations in the above described 
 scenarios~\cite{esterle2022deep}. 
Several studies (e.g.,~\cite{DBLP:journals/tit/GhoshCYR22,DBLP:conf/acsos/DominiAFVE24,DBLP:conf/ecai/Li0TWXZ23}) 
 have attempted to tackle this scenario by proposing algorithms that cluster 
 clients based on similarity metrics, 
 under the assumption that clients within the same cluster have 
 IID data while clusters themselves exhibit Non-IID properties.
Nevertheless, the lack of standardized benchmarks for evaluating such approaches remains 
 a significant limitation. 
Existing benchmarks--like~\cite{DBLP:conf/icde/LiDCH22,DBLP:journals/pami/HuangYSWLDY24}--often rely on synthetic data splits or 
 arbitrary partitioning schemes that fail to capture the realistic geographic
 clustering observed in practice.

To bridge this gap, we introduce \emph{ProFed}\footnote{\url{https://github.com/davidedomini/ProFed}}, 
 a novel benchmark designed specifically
 for proximity-based Non-IID \ac{fl} providing a more realistic and complete evaluation setting.
ProFed leverages well-known computer vision datasets from PyTorch~\cite{DBLP:conf/asplos/AnselYHGJVBBBBC24} 
 and TorchVision~\cite{TorchVision}--like MNIST~\cite{lecun2010mnist}, 
 CIFAR10~\cite{cifar10} and CIFAR100~\cite{cifar100}--and incorporates established data partitioning methods from the literature, 
 such as Dirichlet distribution-based splits~\cite{DBLP:conf/nips/LinKSJ20,DBLP:conf/nips/WangLLJP20}. 
Moreover, by enabling researchers to control the degree of data skewness, 
 this approach allows for fine-grained experimentation and analysis. 
%

To demonstrate the utility and usability of ProFed, 
 we conducted experiments using three state-of-the-art algorithms, namely:
 FedAvg~\cite{DBLP:conf/aistats/McMahanMRHA17}, 
 FedProx~\cite{DBLP:conf/mlsys/LiSZSTS20} and 
 Scaffold~\cite{DBLP:conf/icml/KarimireddyKMRS20}.

The remainder of this paper is organized as follows: 
 \Cref{sec:background} provides a background and review of related work. 
 \Cref{sec:formalization} formalizes the problem statement. 
 \Cref{sec:benchmark} details the structure and implementation of the proposed benchmark. 
 \Cref{sec:evaluation} presents an evaluation of our benchmark running experiments with multiple algorithms. 
 Finally, \Cref{sec:future} concludes the paper and outlines paths for future research.

\section{Background and Related Work}\label{sec:background}

\subsection{Federated Learning}

\Acl{fl}~\cite{DBLP:conf/aistats/McMahanMRHA17} is a machine learning framework
 that aims at collaboratively train a joint global model from
 distributed datasets. 
This technique has been introduced to enable the training of 
 models without the need of collecting and merging multiple datasets into a 
 single central server, thus making it possible to work in contexts with strong 
 privacy concerns---for instance, 
 hospitals~\cite{DBLP:journals/csur/NguyenPPDSLDH23,DBLP:journals/toit/PfitznerSA21}
 or banks~\cite{DBLP:series/lncs/LongT0Z20,DBLP:conf/bigdata2/YangZYL019}. 
Moreover, in highly distributed systems, given the large volume of generated data,
 it becomes inconvenient, or also infeasible, to move all the data to a 
 central server~\cite{DBLP:journals/comsur/NguyenDPSLP21}.

\begin{figure}[htb]
    \centering
    \includegraphics[width=0.8\columnwidth]{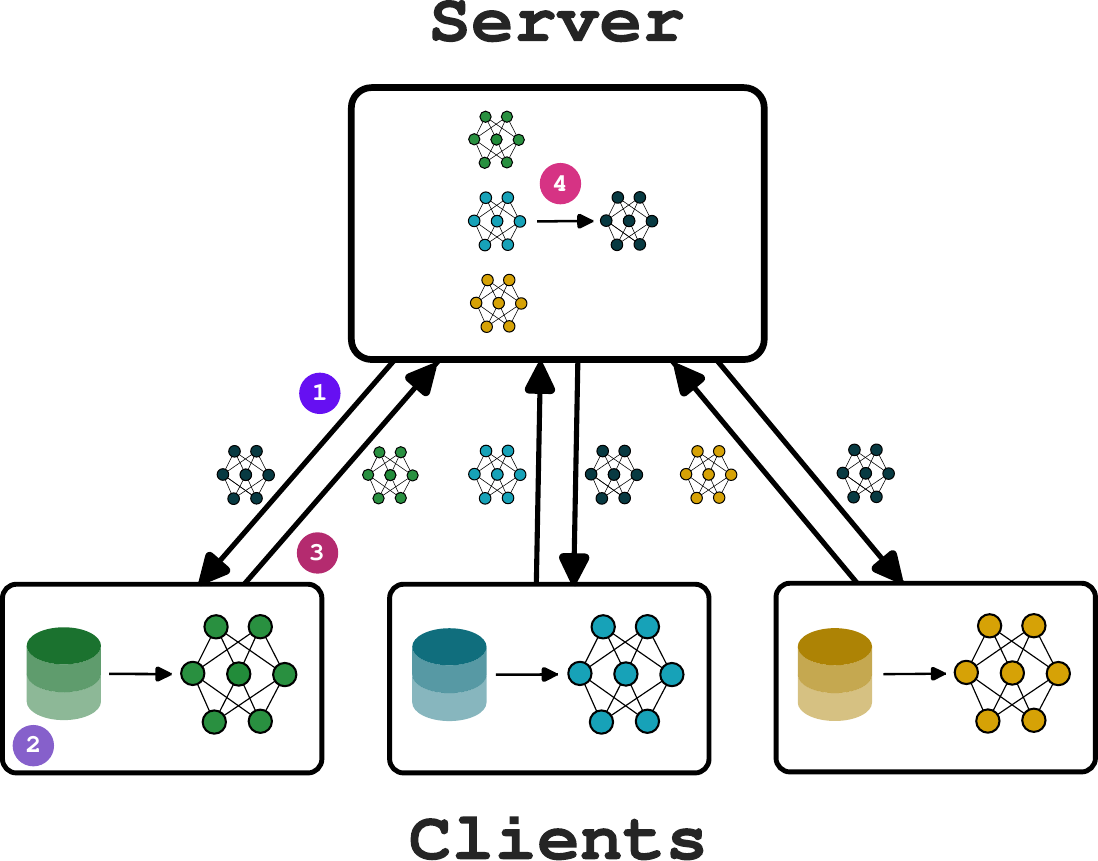}
    \caption{A graphical representation of the Federated Learning process. 
    The numbers (1-4) present the steps of the learning flow discussed above.}
    \label{fig:fl}
\end{figure}

In the literature, most \ac{fl} algorithms share a common learning flow (see~\Cref{fig:fl}):
\begin{enumerate}
    \item \emph{Model initialization}: the central server initializes a common base model 
     that is shared with each client;
    \item \emph{Local Learning}: each client performs one or more steps
     of local learning on its own dataset;
    \item \emph{Local models sharing}: each client sends back to the
     central server the new model trained on its own data
     (i.e. the local model);
    \item \emph{Local models aggregation}: the local models collected
     by the central server are aggregated to obtain the new
     global model.
\end{enumerate}
This process is carried out iteratively for a predefined number of global rounds.
The most common and simple models aggregation method is called FedAvg; it was 
 introduced in~\cite{DBLP:conf/aistats/McMahanMRHA17} and consists in performing 
 an average of the local models to obtain the next global model.

The classical client-server architecture presents some limitations, namely:
\begin{enumerate*}[label=(\roman*)]    
    \item in federations with many clients, the server may be a bottleneck;
    \item the server is a single point of failure, and if it fails to 
     communicate with the clients, the entire the learning process is 
     interrupted; and
    \item the server must be a trusted entity, which could be a challenging 
     constraint in some scenarios.
\end{enumerate*}
For these reasons, despite the commonly adopted learning flow, in recent years 
 alternative algorithms based on peer-to-peer or hierarchical networks 
 have also been proposed~\cite{DBLP:journals/jpdc/HegedusDJ21,DBLP:conf/dsn/WinkN21,DBLP:conf/icc/Liu0SL20}.

\subsection{Learning from Heterogeneous Data}
Federated learning algorithms achieve remarkable performance when data are 
 homogeneously distributed among clients~\cite{DBLP:conf/middleware/NilssonSUGJ18}. 
However, in real-life datasets, data are often heterogeneous, 
 namely \emph{non-independently and identically distributed} (Non-IID). 
The skewness of data can be categorized in various ways based on how they 
 are distributed among the clients~\cite{DBLP:conf/icde/LiDCH22,DBLP:journals/ftml/KairouzMABBBBCC21}. 
The main categories include:
\begin{enumerate*}[label=(\roman*)]    
    \item \emph{feature skew}: all clients have the same labels but different feature 
     distributions (e.g., in a handwritten text classification task, we may have the same letters 
     written in different calligraphic styles);
    \item \emph{label skew}: each client has only a subset of the classes; and
    \item \emph{quantity skew}: each client has a significantly different amount of data compared to others.
\end{enumerate*}
In this work, we focus on label skew. 
More in detail, given the features $x$, a label $y$, and 
 the local distribution of a device $i$, denoted as:
\begin{equation}
    \mathbb{P}_i(x,y) = \mathbb{P}_i(x | y) \cdot \mathbb{P}_i(y)
\end{equation}
the concept of label skew, between two distinct devices $k$ and $j$, is defined as follows~\cite{DBLP:journals/pami/HuangYSWLDY24}:
\begin{equation}
    \begin{split}
    \mathbb{P}_k(y) &\neq \mathbb{P}_j(y) \\
    \mathbb{P}_k(x | y) &= \mathbb{P}_j(x | y)
    \end{split}
\end{equation}

In other words, different clients have distinct label distributions (i.e., $\mathbb{P}_i(y)$) 
 while maintaining the same underlying feature distribution for each label (i.e., $\mathbb{P}_i(x|y)$).

To tackle this problem, several algorithms have been proposed in the literature.
FedProx~\cite{DBLP:conf/mlsys/LiSZSTS20} tries to improve FedAvg adding a regularization
 term to the loss function to limit the size of updates, while Scaffold~\cite{DBLP:conf/icml/KarimireddyKMRS20}
 introduces control variate to limit the drift of local updates. 
All these algorithms do not make assumptions about how data is 
 non-independently and identically distributed among clients. 
For this reason, during evaluation, well-known datasets are used and split following non-homogeneous 
 distributions (e.g., Dirichlet distribution)---like in~\cite{DBLP:journals/corr/abs-2302-02949,DBLP:conf/icde/LiDCH22}.

Though this heterogeneity can be caused by various factors~\cite{DBLP:journals/ijon/ZhuXLJ21}, 
 it frequently arises from the spatial distribution of devices, partially related to the problem of cross-client shift.
This builds on the assumption that devices in \emph{spatial proximity} have similar experiences and 
 make similar observations~\cite{esterle2022deep}, 
 as the phenomena to capture is \emph{intrinsically} context dependent.
As an example, consider the urban traffic prediction scenario: traffic patterns observed by 
 devices within the same city district are likely to be more similar compared to those observed
 by devices in different districts.
Therefore, various algorithms based on client clustering, with the assumption that clients can be grouped such that data are IID, have been proposed.
Some examples are: PANM~\cite{DBLP:journals/tbd/LiLLZSLZWW24}, FedSKA~\cite{DBLP:conf/ecai/Li0TWXZ23}, 
 FBFL~\cite{DBLP:conf/coordination/DominiAEV24}, PBFL~\cite{DBLP:conf/acsos/DominiAFVE24}, 
 and IFCA~\cite{DBLP:journals/tit/GhoshCYR22}.
 
\subsection{Related Benchmark and Motivation}
Over the years, several benchmarks have been proposed for federated learning, 
 typically focusing on standard datasets split homogeneously across multiple clients, 
 such as~\cite{DBLP:journals/corr/abs-2007-13518,DBLP:journals/corr/abs-2007-14390}. 
However, recent work has shifted its focus towards addressing various data shifts. 
For instance, FedScale~\cite{DBLP:conf/icml/LaiDSLZMC22} offers a comprehensive platform for
 evaluating multiple aspects of federated learning at scale, including system 
 efficiency, statistical efficiency, privacy, and security.
FedScale incorporates a diverse set of realistic datasets and takes into account client resource constraints.

Similarly, LEAF~\cite{DBLP:journals/corr/abs-1812-01097}, another relevant framework, emphasizes reproducibility 
 through its open-source datasets, metrics, and reference implementations. 
LEAF provides granular metrics that assess not only model performance but also the computational and 
 communication costs associated with training in federated settings. 
Additionally, LEAF supports multiple configurations, enabling users to explore 
 different facets of federated learning.

\paragraph{Motivation}
Despite the proposed benchmarks are already valuable resources for the research community, 
 they do not consider one aspect that is crucial in real-world scenarios: 
 \emph{the spatial distribution of devices}.
In fact, in many applications, devices are geographically distributed, and the data they collect 
 is often correlated with their location.
This is where ProFed comes into play, providing a benchmark that simulates data splits 
 with varying degrees of skewness across different regions, thus enabling researchers to evaluate 
 federated learning algorithms in a more realistic and complete setting.

\section{Benchmark description}\label{sec:description}
\subsection{Formalization}\label{sec:formalization}
\begin{figure}[htb]
    \centering
    \includegraphics[width=0.6\columnwidth]{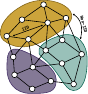}
    \caption{A visual representation of data skewness between different subregions.
    Each color represents a different data distribution.
    It is possible to see how, inside the same subregion data are homogeneously distributed,
    while between different subregion data exhibit Non-IID properties.
    }
    \label{fig:areas}
\end{figure}

As depicted in~\Cref{fig:areas},
 we consider a spatial area $A = \{a_1, \dots, a_k\}$ divided into $k$ distinct 
 contiguous subregions.
Each subregion $a_j$ has a unique data distribution $\Theta_j$ and provides specific 
 localized information.
This means that, given two regions $i,j$ and the respective data distributions
 $\Theta_i$ and $\Theta_j$, a sample $d'$ from $\Theta_i$ is distinctively
 dissimilar from a sample $d''$ from $\Theta_j$ (namely, the data is non-IID).
Whereas, giving two data distributions 
$\Theta_i$ and $\Theta_j$ from the same region $i$, $d'$ and $d''$ sample from same $\Theta_i$, their 
 difference $m(d',d'')$ is negligible (namely, the data is homogeneous).

This dissimilarity can be quantified using a specific distance metric $m(d', d'')$, 
 which determines the disparity between two distributions.
Formally, given an error bound $\delta$, the dissimilarity intra-region and 
 inter-region can be quantified as follows:
\begin{equation}
    \forall i\neq j, \forall d,d' \in \Theta_i, \forall d'' \in \Theta_j: m(d, d') \leq \delta < m(d, d'')
\end{equation}

In $A$, a set of \emph{sensor nodes} $S = \{s_1, \dots, s_n \}$ ($n \gg |A| $) 
 are deployed---for instance, these sensor nodes may be smartphones or cameras in cars.
Each sensor node is assumed to be capable of processing data and to have 
 enough computational power to be able to participate in the 
 federated learning process.
Locally, each node $i$ creates a dataset $D_i$ of samples perceived from the 
 data distribution $\Theta_j$ of its respective region $j$.  
In this work, we consider a general classification task where each sample $d$ in the
 data distribution $\Theta_j$ consists of a feature vector $x$ and a label $y$.
Therefore, the complete local dataset $D_i$ is represented as 
 $D_i = \{ (x_1, y_1), \dots, (x_m, y_m) \}$.

\subsection{Implementation Details}\label{sec:benchmark}
ProFed implementation is based on PyTorch~\cite{DBLP:conf/asplos/AnselYHGJVBBBBC24} 
 and TorchVision~\cite{TorchVision}, 
 as it has been specifically designed to facilitate and 
 standardize research experiments within the scenario described in~\Cref{sec:formalization}. 
In the following, we detail the implemented methods to synthesize skewed datasets, 
 the supported datasets and the API of the benchmark.

\begin{figure*}
\centering
\begin{subfigure}{0.3\textwidth}
    \includegraphics[width=\textwidth]{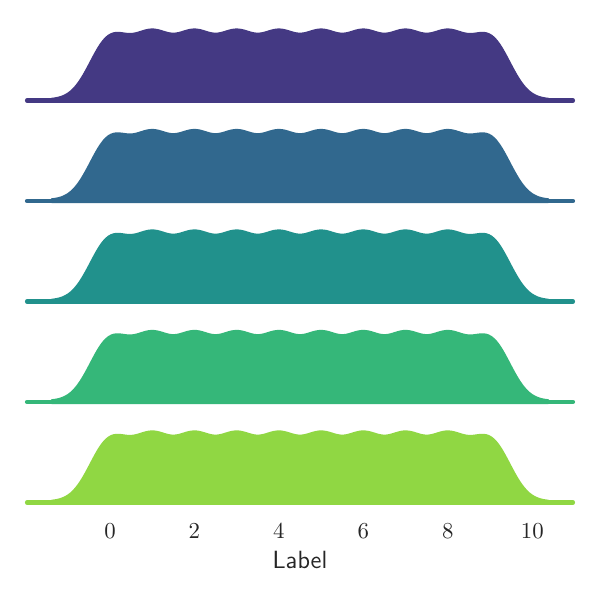}
    \caption{IID data.}
    \label{fig:iid}
    \end{subfigure}
\begin{subfigure}{0.3\textwidth}
    \includegraphics[width=\textwidth]{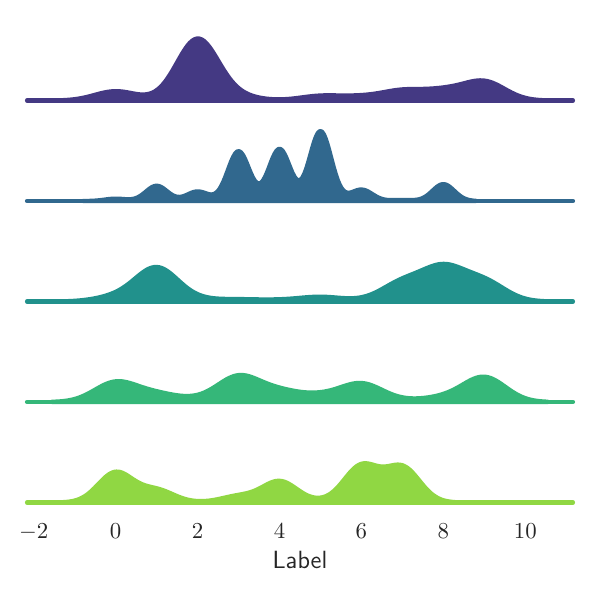}
    \caption{Non-IID data (Dirichlet distribution).}
    \label{fig:dirichlet}
    \end{subfigure}
\begin{subfigure}{0.3\textwidth}
    \includegraphics[width=\textwidth]{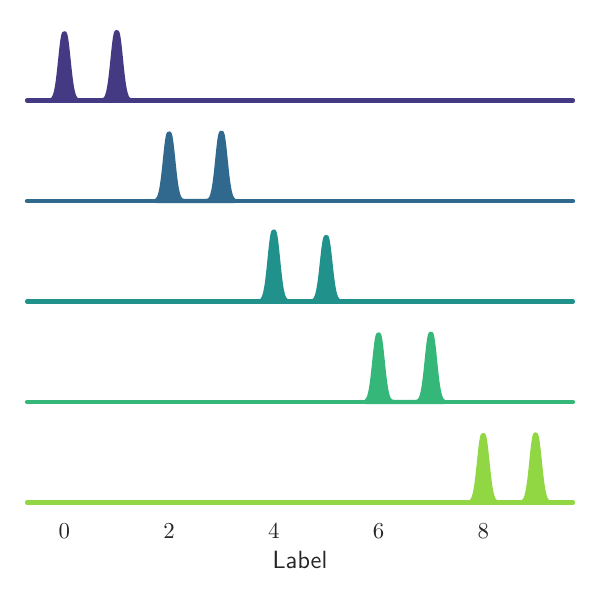}
    \caption{Non-IID data (hard partitioning).}
    \label{fig:hard}
\end{subfigure}
\caption{ 
    A graphical representation of three different data distribution in $5$ subregions.
    Each color represents a different subregion.
 }
\label{fig:skewness}
\end{figure*}

\paragraph{Data distribution}
As part of our analysis, 
 we reviewed several studies in the literature on non-IID 
 Federated Learning to identify and select the most commonly used partitioning methods.
We observed that several works employed \emph{Dirichlet} distribution for data partitioning. 
This approach results in each party having instances of most labels, 
 although the distribution is highly imbalanced, 
 with some labels being underrepresented and others heavily overrepresented.
The degree of skewness in the distribution can be adjusted using the parameter 
 $\alpha$, where lower values of $\alpha$ result in a more skewed distribution.
In the literature, $\alpha$ values typically range from 0.1 to 1.0, 
 with lower values producing more extreme non-IID distributions and higher values 
 resulting in more balanced ones.
A value of $\alpha=0.5$ is commonly used as it provides a moderate level of 
 data heterogeneity suitable for evaluating FL algorithms.
An example of this distribution, for $5$ subregions and the MNIST dataset (with $10$ classes), is represented in~\Cref{fig:dirichlet}.
The second data distribution considered in this work is referred to as \emph{hard},
 where each party has access to only a subset of labels.
This leads to a significantly more skewed distribution, 
 making it considerably more challenging for the stability of learning algorithms.
An example of this distribution, applied to the Dirichlet scenario, is represented in~\Cref{fig:hard}.
In ProFed, we enable fine-grained control over the data distribution, allowing either 
 balanced label subsets across regions or customizable cardinality per subregion.
For comparative analysis, we also implement an IID split as a baseline distribution (\Cref{fig:iid}).
While existing approaches in the literature typically apply these partitioning methods 
 directly at the device level (e.g.,~\cite{DBLP:journals/corr/abs-2302-02949,DBLP:conf/nips/LinKSJ20}), 
 our framework introduces an intermediate layer of regional clustering.
Indeed, ProFed first distributes data heterogeneously among subregions and 
 then splits them homogeneously among the devices within the same subregion, 
 thereby creating clustered heterogeneity---see~\Cref{fig:areas}.
Finally, to support extensibility, ProFed enables custom partitioning of 
 the dataset based on a distribution specified by the user. 
This distribution is represented as a matrix with $N$ columns 
 (corresponding to the number of labels in the dataset) 
 and $M$ rows (corresponding to the number of subregions). 
Each cell $(i,j)$ in the matrix indicates the proportion of instances with label $i$ 
 that should be assigned to subregion $j$.

\paragraph{Supported datasets}

\begin{figure}
    \centering
    \begin{subfigure}{\columnwidth}
        \includegraphics[width=\textwidth]{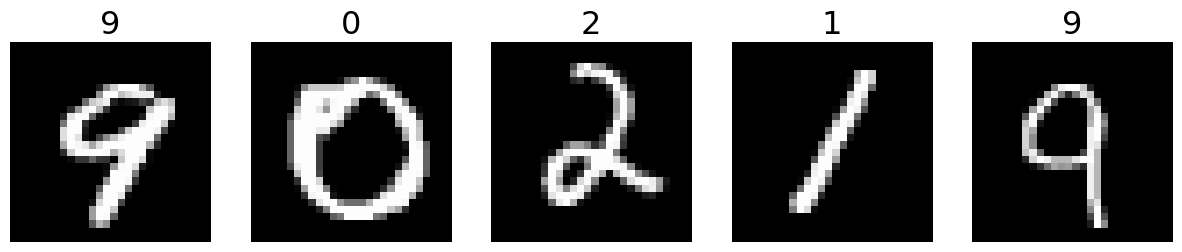}
        \caption{Data sample from MNIST dataset.}
        \label{fig:mnist}
    \end{subfigure}
    \begin{subfigure}{\columnwidth}
        \includegraphics[width=\textwidth]{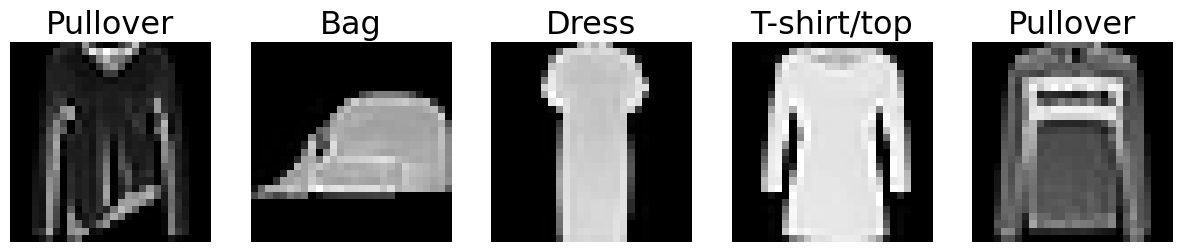}
        \caption{Data sample from Fashion MNIST dataset.}
        \label{fig:fashion}
    \end{subfigure}
    \begin{subfigure}{\columnwidth}
        \includegraphics[width=\textwidth]{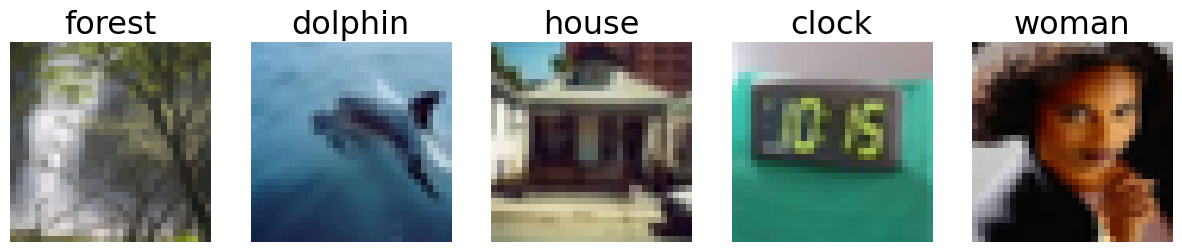}
        \caption{Data sample from CIFAR100 dataset.}
        \label{fig:cifar100}
    \end{subfigure}
    \caption{Overview of the supported datasets. }
    \label{fig:dataset}
\end{figure}

ProFed supports a variety of datasets from TorchVision, 
 all of which are widely recognized in computer vision literature 
 (a summary of each dataset details is given in~\cref{tab:datasets}). 
First, we included a frequently used dataset in research as a baseline 
 for performance comparison, namely: MNIST~\cite{lecun2010mnist}.
It consists of grayscale $28 \times 28$ pixel images of handwritten 
 digits across 10 classes---a random sample of images is shown in~\Cref{fig:mnist}. 
ProFed also includes two notable extensions of MNIST: 
 Fashion MNIST~\cite{DBLP:journals/corr/abs-1708-07747} and Extended MNIST~\cite{DBLP:journals/corr/CohenATS17}. 
Fashion MNIST, introduced by Zalando, features 28x28 grayscale images of clothing items, 
 also categorized into 10 classes 
 (e.g., ``Trouser'', ``Pullover'', ``Dress'' and many more)---see~\Cref{fig:fashion}. 
Extended MNIST, on the other hand, contains 28x28 images of handwritten Latin alphabet letters, 
 spanning 27 classes, and is considered more challenging than the previous two. 
Additionally, ProFed supports CIFAR10 and CIFAR100, which provide colorized 32x32 
 images (3-channel RGB)---see~\Cref{fig:cifar100}. 
CIFAR10 includes 10 classes, while CIFAR100 expands to 100 classes, offering greater complexity. 
Notably, all these datasets maintain balanced class distributions; 
 for instance, CIFAR10 provides 6,000 training instances per class.

\begin{table*}[t]
    \centering
    \caption{Summary of the characteristics of the supported datasets in the benchmark.}
    \label{tab:datasets}
    \begin{tabular}{lrrrr}
    \toprule
    \textbf{Dataset} & \textbf{Training Size} & \textbf{Test Size} & \textbf{Features} & \textbf{Classes} \\ 
    \midrule
    MNIST           & 60,000 & 10,000 & 784  & 10  \\ 
    Fashion MNIST   & 60,000 & 10,000 & 784  & 10  \\ 
    Extended MNIST  & 124,800 & 20,800 & 784  & 27  \\ 
    CIFAR-10       & 50,000 & 10,000 & 3,072 & 10  \\ 
    CIFAR-100      & 50,000 & 10,000 & 3,072 & 100 \\ 
    \bottomrule
    \end{tabular}
\end{table*}

\paragraph{Benchmark API}

ProFed has been designed with usability and ergonomics in mind. 
To achieve this, its API provides all the necessary methods 
 to manage the referenced use case seamlessly---an example is provided in Listing~\ref{lst:api-example}.
First, ProFed allows users to download the selected dataset directly 
 from TorchVision and automatically generate training and validation subsets.
Second, and most important, given a dataset and a predefined number of subregions,
 it enables users to distribute data among subregions following the
 specified distribution strategy.
Finally, once the data distribution among subregions is established, 
 ProFed facilitates the creation of datasets for individual devices. 
At this stage, users can either choose a homogeneous distribution, 
 where a fixed number of devices is assigned to each subregion, or define 
 the exact number of devices per subregion, thereby enhancing flexibility 
 in the configuration.
Each device-specific dataset is represented as an instance of the \texttt{Subset} class from PyTorch, 
 ensuring full compatibility with existing learning algorithms based on it.

\noindent\begin{minipage}{\columnwidth}
\lstdefinestyle{proFedStyle}{
    backgroundcolor=\color{gray!10},
    frame=single,
    rulecolor=\color{blue!60},
    frameround=tttt,
    framesep=5pt,
    breaklines=true,
    captionpos=b,
    numberstyle=\tiny\color{gray}
}

\begin{lstlisting}[style=proFedStyle, caption={Example of usage of ProFed to partition the MNIST dataset between $5$ subregions and $50$ devices.}, label={lst:api-example}]
from ProFed.partitionings import Partitioner

partitioner = Partitioner()
dataset = partitioner.download_dataset('MNIST')
training_set, validation_set = partitioner.train_validation_split(dataset, 0.8)
partitioning_names = ['IID', 'Dirichlet', 'Hard']
areas = 5    

for name in partitioning_names:
    partitioning = partitioner.partition(name, training_set, areas)
    devices_data = partitioner.subregions_distributions_to_devices_distributions(
        partitioning,
        number_of_devices=50
    )
    ...
\end{lstlisting}
\end{minipage}

\section{Experiments}\label{sec:evaluation}

\subsection{Experimental setup}
To evaluate the effectiveness and the usability of ProFed, 
 we performed several experiments using the supported datasets and three
 state-of-the-art learning algorithms, namely: 
 FedAvg~\cite{DBLP:conf/aistats/McMahanMRHA17}, 
 FedProx~\cite{DBLP:conf/mlsys/LiSZSTS20} and 
 Scaffold~\cite{DBLP:conf/icml/KarimireddyKMRS20}.
Data have been synthetically split to obtain multiple non-IID distributions using all 
 the supported methods, namely: IID, Dirichlet (with $\alpha=0.5$) and Hard.
In particular, we conducted several experiments with a variable number of areas $A \in \{3, 6, 9\}$.
All the experiments were implemented using PyTorch.
The same hyperparameters were employed for all the approaches. 
A simple MLP neural network with 128 neurons in the hidden layer was
 trained for a total of $30$ global rounds. 
At each global round, each device performed $2$ epochs of local learning 
 using a batch size of $32$, the ADAM optimizer with a learning rate
 of $0.001$, and a weight decay of $0.0001$.
All the experiments have been repeated multiple time with varying seed ($5$ different seeds), 
 to produce stronger results. 
Thus, the total number of experiments is $120$.

All the code is distributed with a permissive license and publicly available for
 reproducibility on GitHub.

\subsection{Discussion}
The results of our experiments were systematically collected 
 during training, validation and testing phases.

\begin{figure*}
    \centering


    \begin{subfigure}{0.32\textwidth}
        \includegraphics[width=\textwidth]{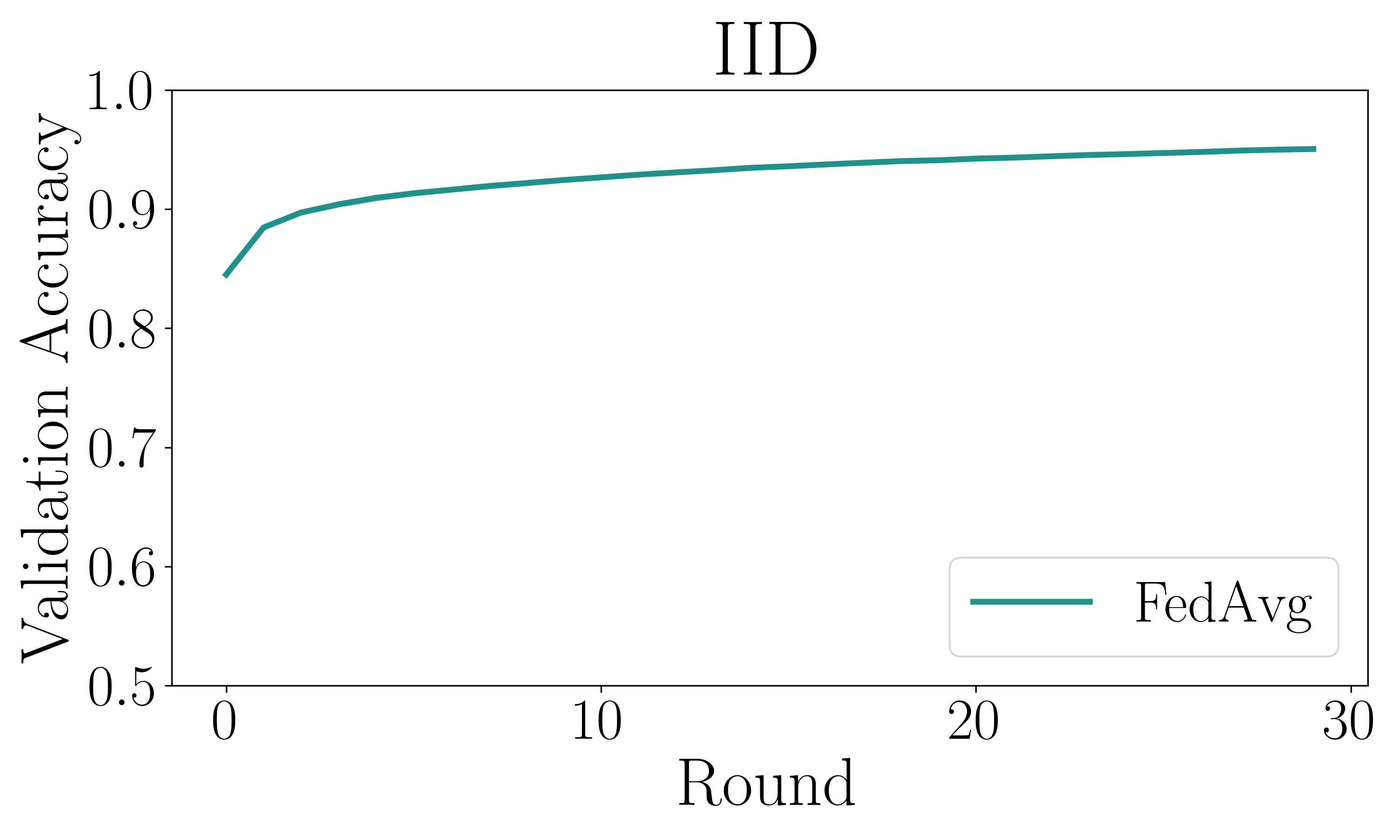}
    \end{subfigure}
    \begin{subfigure}{0.32\textwidth}
        \includegraphics[width=\textwidth]{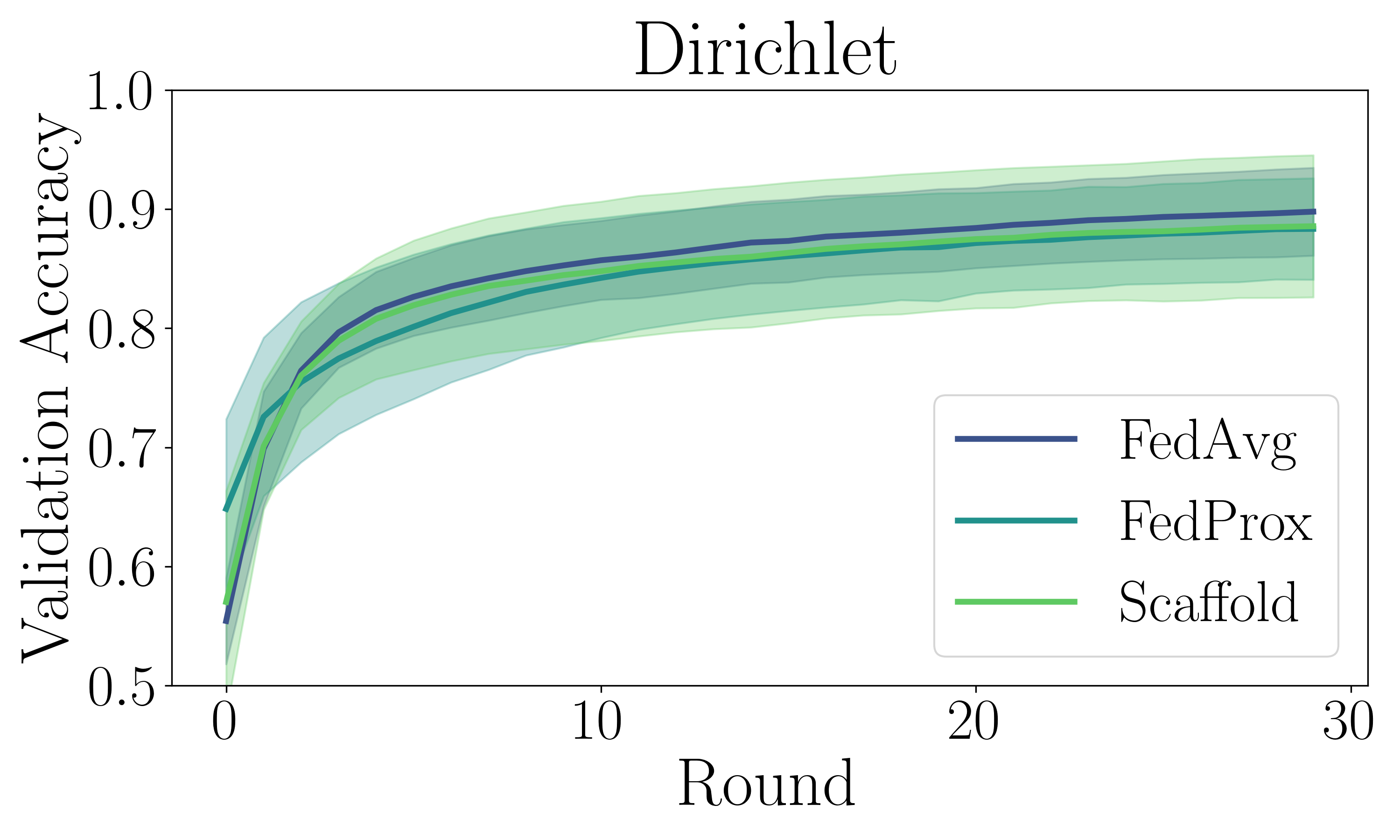}
    \end{subfigure}
    \begin{subfigure}{0.32\textwidth}
        \includegraphics[width=\textwidth]{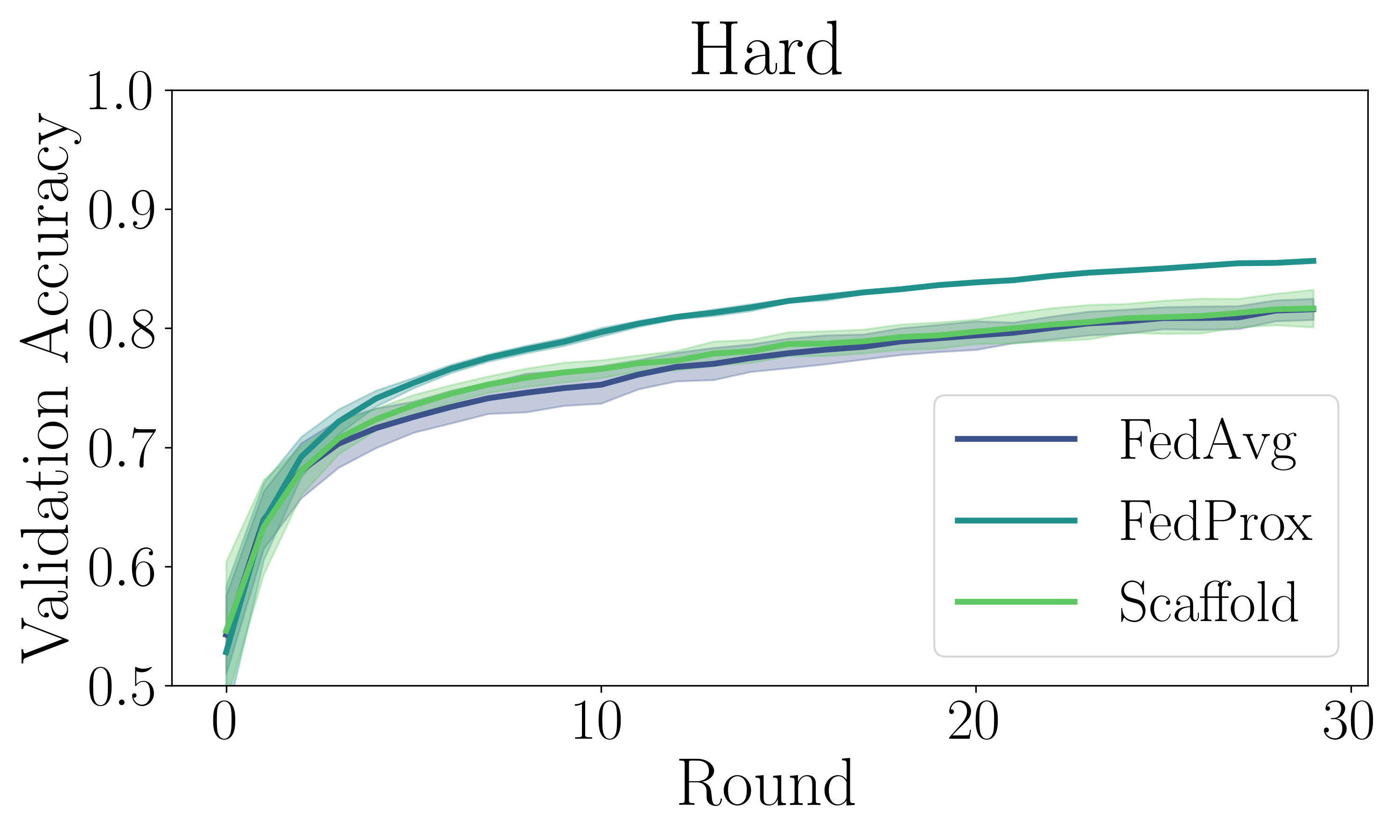}
    \end{subfigure}


    \begin{subfigure}{0.32\textwidth}
        \includegraphics[width=\textwidth]{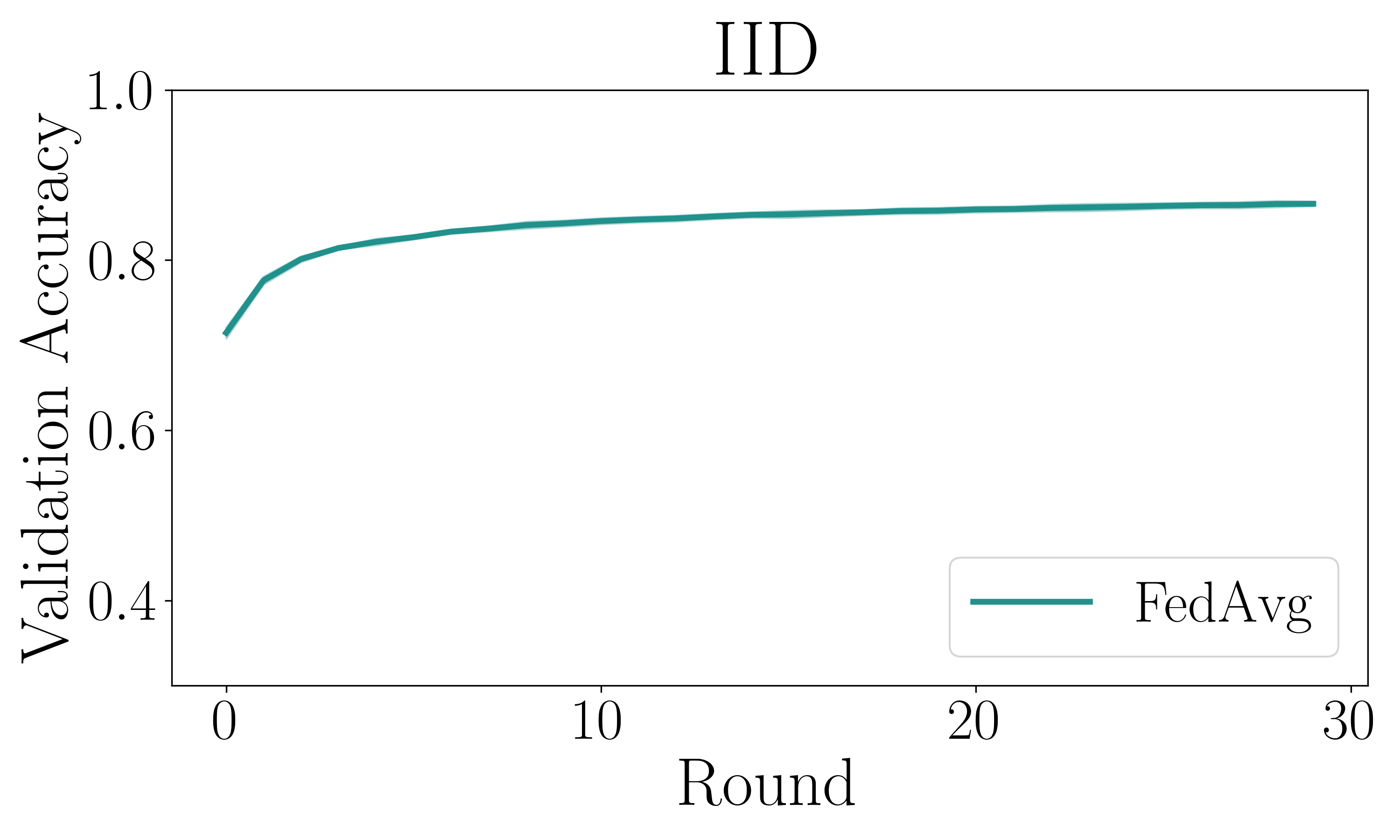}
    \end{subfigure}
    \begin{subfigure}{0.32\textwidth}
        \includegraphics[width=\textwidth]{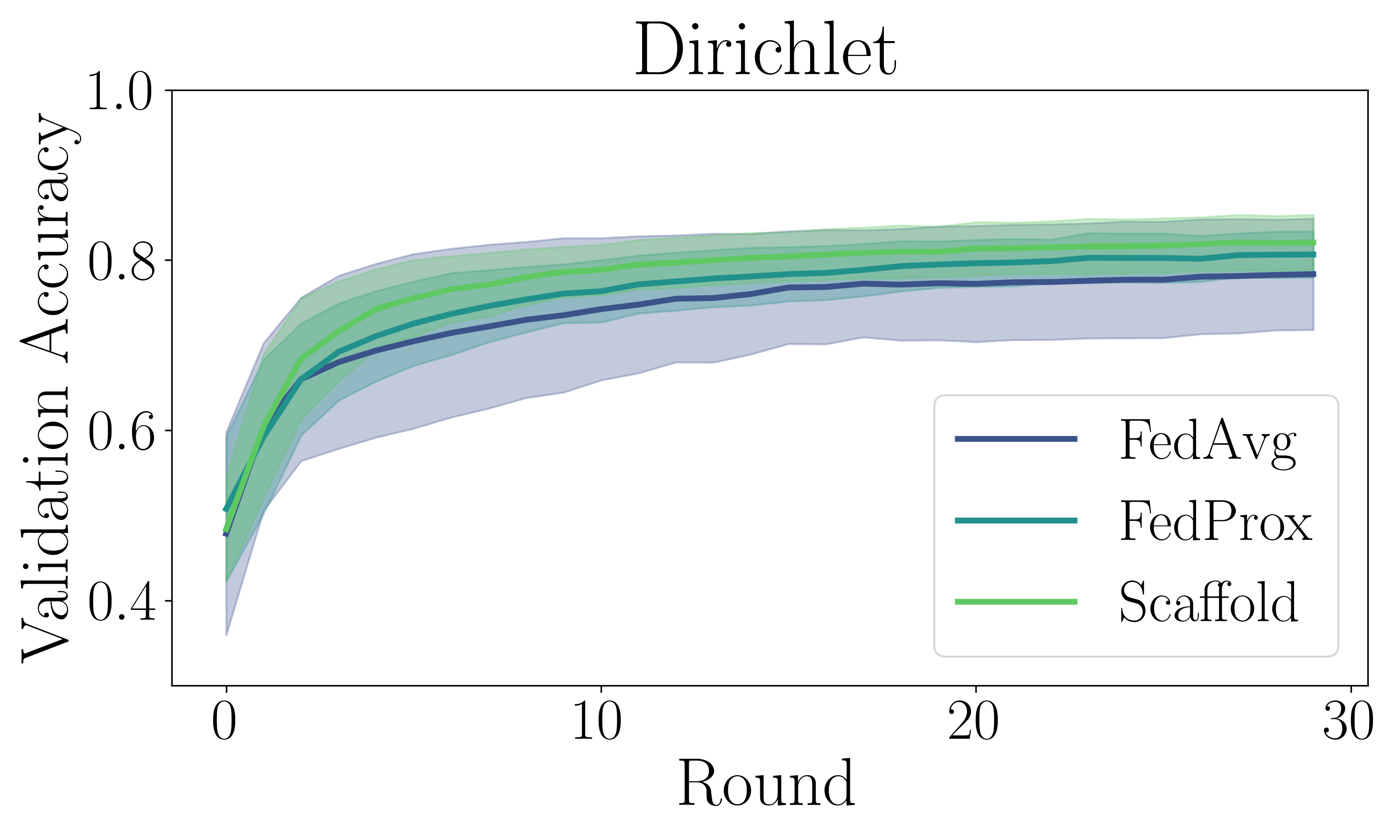}
    \end{subfigure}
    \begin{subfigure}{0.32\textwidth}
        \includegraphics[width=\textwidth]{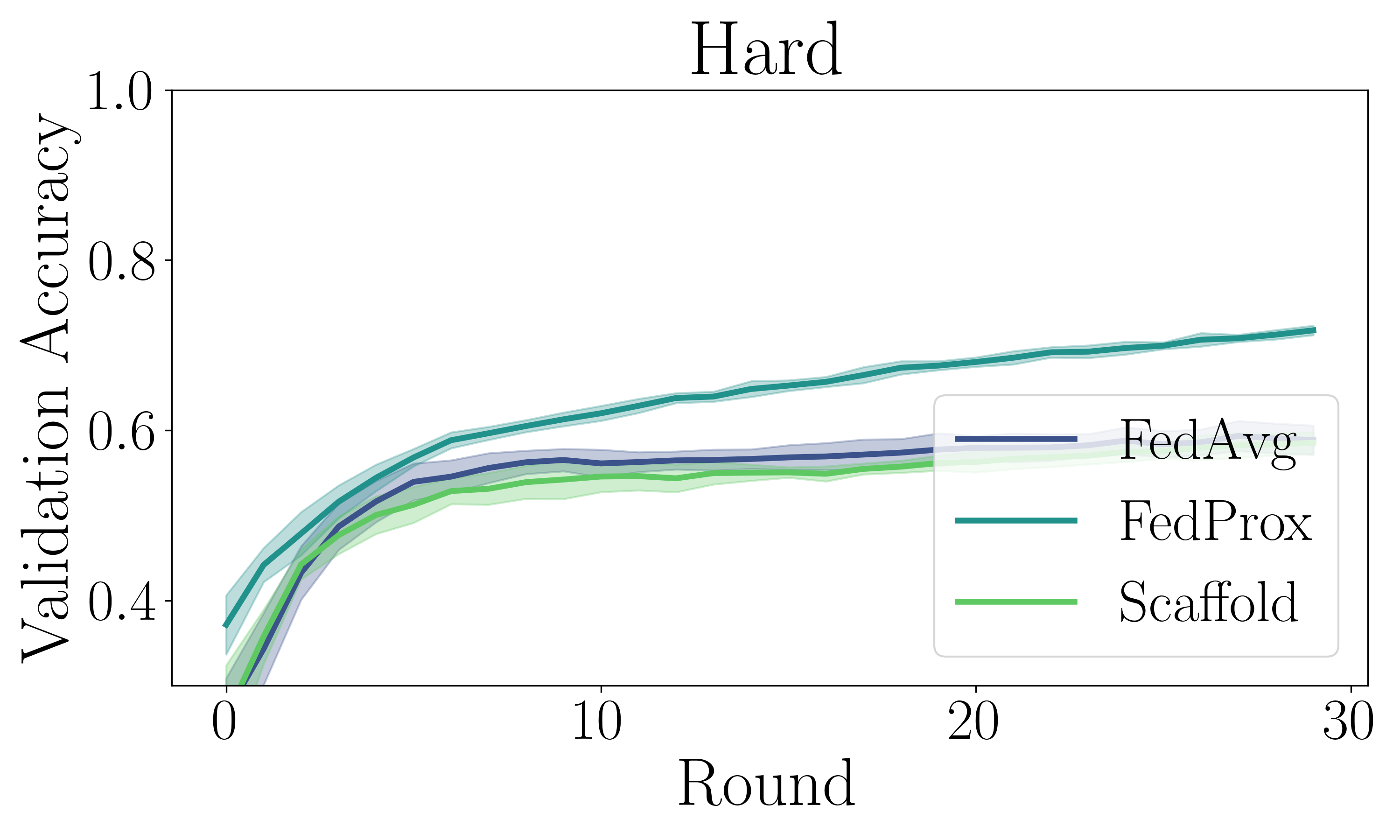}
    \end{subfigure}


    \begin{subfigure}{0.32\textwidth}
        \includegraphics[width=\textwidth]{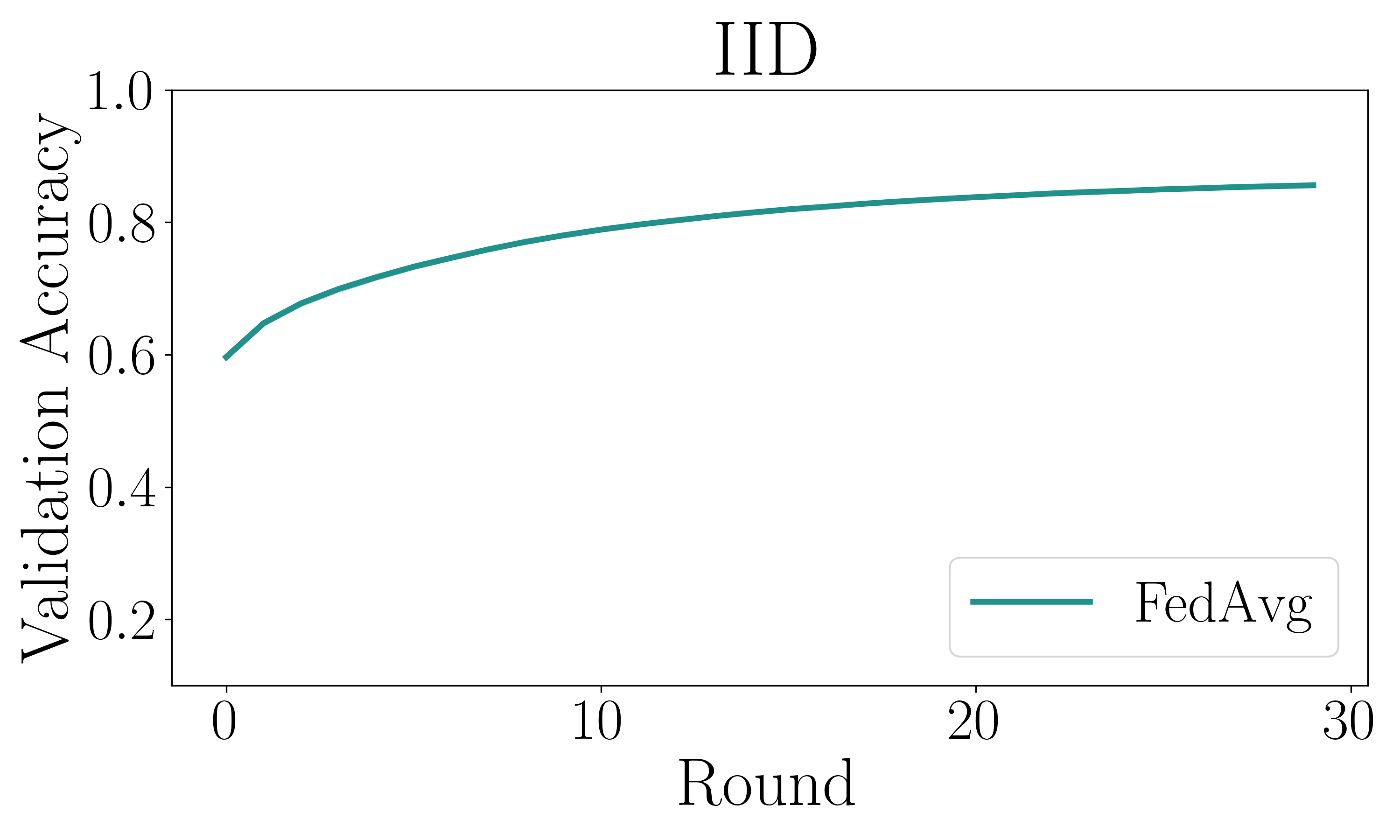}
    \end{subfigure}
    \begin{subfigure}{0.32\textwidth}
        \includegraphics[width=\textwidth]{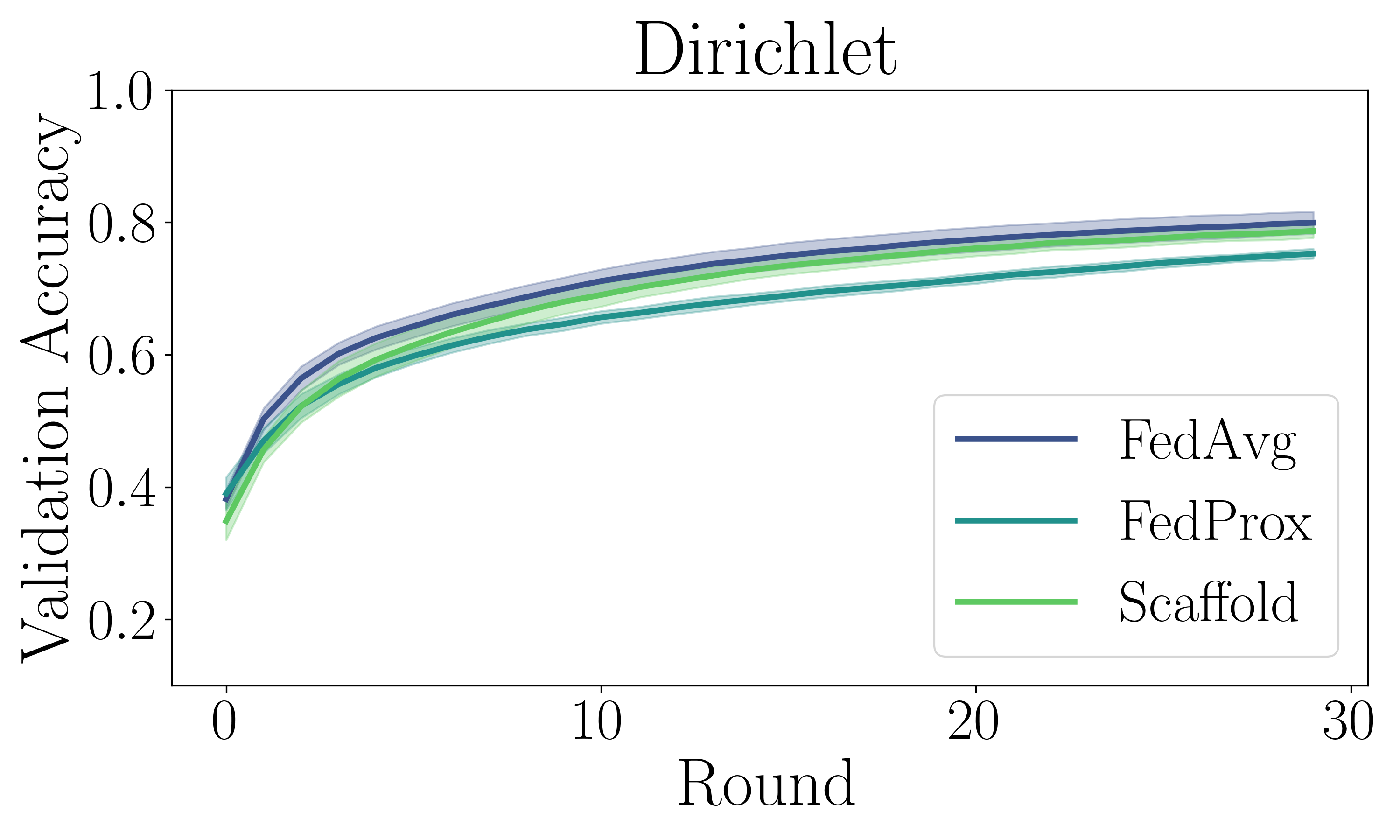}
    \end{subfigure}
    \begin{subfigure}{0.32\textwidth}
        \includegraphics[width=\textwidth]{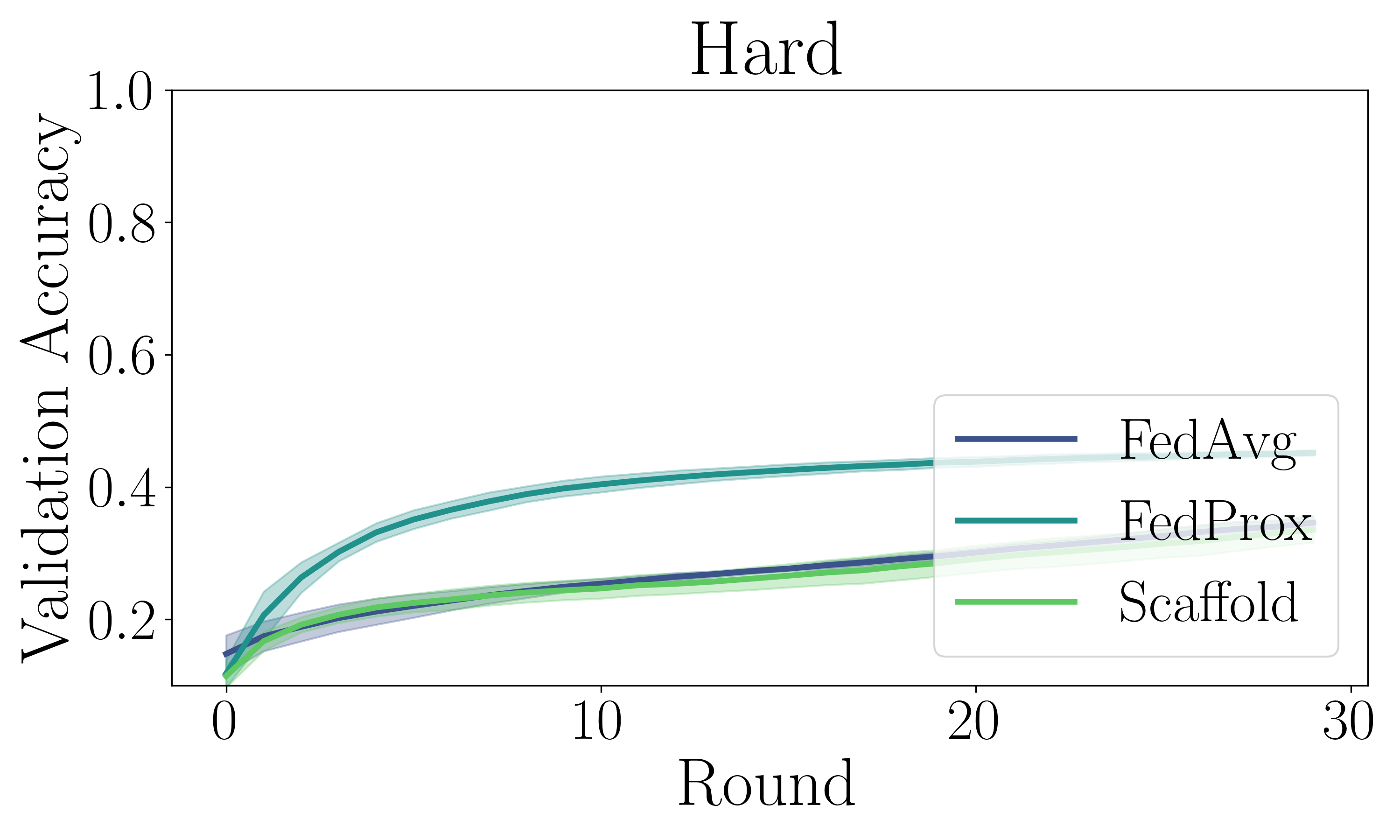}
    \end{subfigure}

\caption{
    A visual representation of validation results. 
    The first row represents results on the MNIST dataset, while second and third rows represents respectively results on 
    FashionMNIST and Extended MNIST datasets.
    The three columns represents the three partitioning methods (i.e., IID, Dirichlet and Hard), 
 }
\label{fig:exp}
\end{figure*}

Initially, we utilized a homogeneous data distribution (i.e., IID) as a baseline to 
 evaluate model stability and accuracy. 
In this phase, we employed FedAvg exclusively, as it is known to perform well under 
 IID conditions, whereas alternative methods such as Scaffold and FedProx 
 are specifically designed to address non-IID scenarios.
Our results indicate that training FedAvg under IID data conditions 
 is stable and yields high accuracies. 
This is evident in both the validation phase and the testing phase.
Validation stability is shown in the first column of~\Cref{fig:exp}
 where accuracy monotonically increase towards the maximum. 
Similarly, test stability is illustrated in~\Cref{tab:test} where the model 
 maintains high accuracy and minimal fluctuations.

The importance of the data distribution is highlighted when transitioning 
 from IID to non-IID conditions.
Indeed, when we introduce heterogeneous data distribution, 
 a significant performance drop is observed. 
None of the evaluated algorithms successfully handle extreme levels of data skewness, 
 leading to reduced model accuracy and stability. 
This is particularly evident under hard partitioning conditions. 
For instance, the last row of~\Cref{fig:exp} presents an example of hard partitioning 
 into nine distinct regions, where the validation accuracy drops from
 $80\%$ in the case of IID data to $50\%$. 
Furthermore, the testing phase results reflect a similar trend. 
While FedAvg under IID conditions achieves stable performance above $95\%,$ 
 transitioning to non-IID partitions using a Dirichlet distribution 
 results in greater instability (as evidenced by the increased variance in model accuracy). 
The performance decline is even more pronounced under hard partitioning, 
 further demonstrating the limitations of existing federated learning methods 
 in handling extreme data heterogeneity.
This a clear indication that the current state-of-the-art algorithms are not 
 well-suited to address the challenges posed by spatially distributed data 
 and that further research is needed to develop more robust solutions.
\begin{table}[t]
    \centering
    \caption{Results on the test set for different algorithms with different partitioning methods.}
    \label{tab:test}
    \begin{tabular}{c|ccc}
    \toprule
    \textbf{Algorithm} & \textbf{IID} & \textbf{Dirichlet} & \textbf{Hard} \\ 
    \midrule
    FedAvg        & 0.95 $\pm$ 0.001 & 0.9 $\pm$ 0.04 & 0.81 $\pm$ 0.01  \\ 
    FedProx       & \ding{55} &  0.886 $\pm$ 0.04 & 0.86 $\pm$ 0.01 \\ 
    Scaffold      & \ding{55} & 0.889 $\pm$ 0.06 &  0.81 $\pm$ 0.01 \\ 
    \bottomrule
    \end{tabular}
\end{table}

\begin{figure}
    \centering


    \begin{subfigure}{0.7\columnwidth}
        \includegraphics[width=\textwidth]{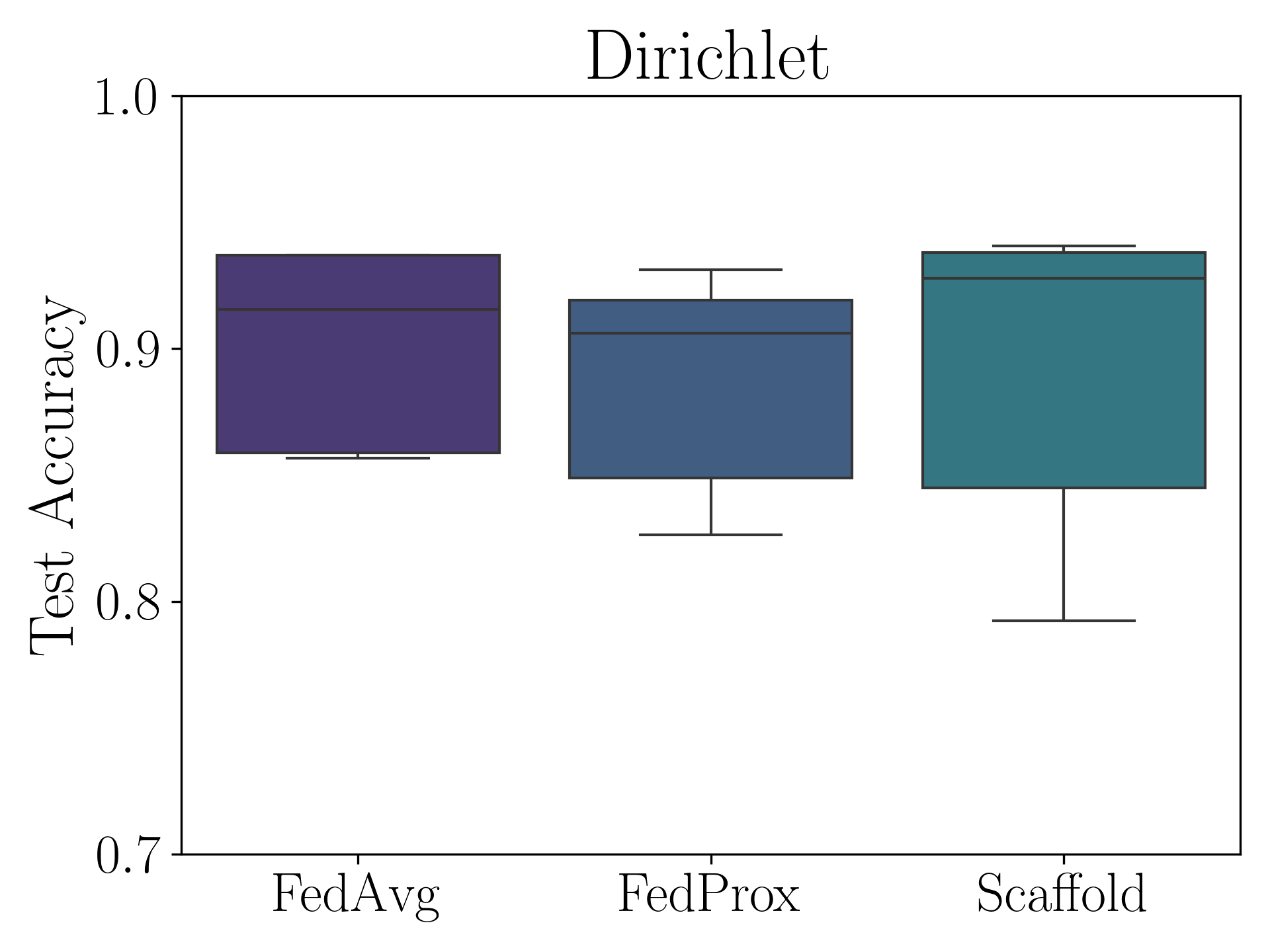}
    \end{subfigure}
    \begin{subfigure}{0.7\columnwidth}
        \includegraphics[width=\textwidth]{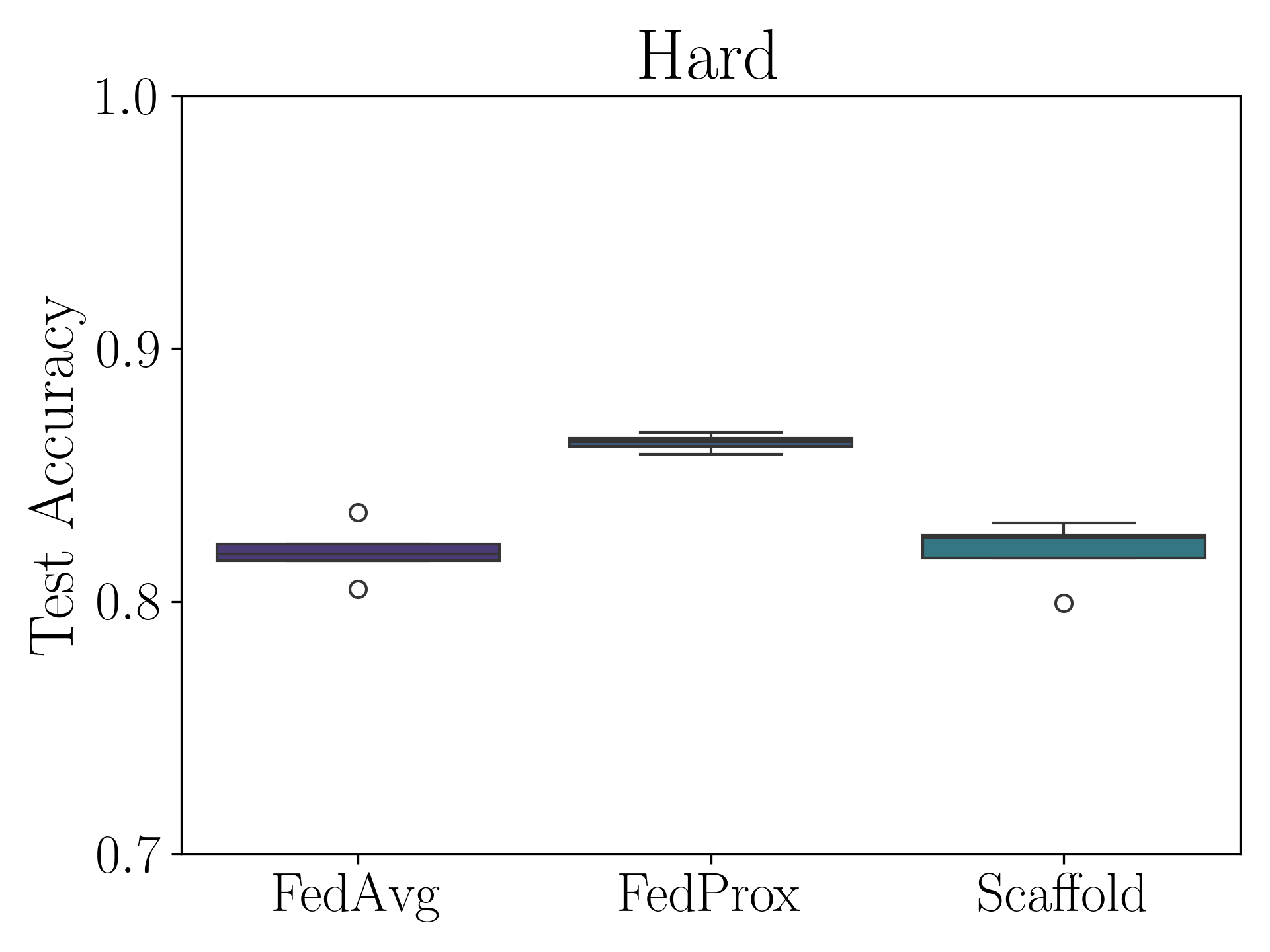}
    \end{subfigure}
\caption{ 
    A visual representation of the test set results for all three implemented algorithms 
     on the MNIST dataset, partitioned into three subregions using both the Dirichlet-based 
     method and hard partitioning.
}
\label{fig:test}
\end{figure}

\section{Conclusion and future work}\label{sec:future}

This work introduces ProFed, a benchmark designed for proximity-based non-independently and 
 identically distributed data in Federated Learning.
While non-IID data has been extensively studied in FL, existing works primarily 
 focus on data skewness at the device level. 
However, prior research suggests that devices in close geographical proximity often share 
 similar environmental conditions, leading to similar data distributions~\cite{esterle2022deep}.
To address this, ProFed introduces the concept of subregions, where data remains non-IID across 
 different regions but is IID within each region. 
This design facilitates the evaluation and comparison of algorithms in Clustered Federated Learning, 
 where devices are grouped into clusters based on data similarity,
 and models are trained accordingly.

We assess the effectiveness and usability of ProFed through experiments with three well-known FL 
 algorithms--FedAvg, FedProx, and Scaffold--using multiple datasets and partitioning 
 strategies from the literature (i.e., IID, Dirichlet, and Hard).

Future research directions could further extend ProFed by incorporating additional 
 tasks (e.g., regression, time-series prediction), 
 modeling heterogeneous computational power across devices, 
 and exploring different types of data skewness---for instance, feature skewness through 
 Gaussian noise~\cite{DBLP:conf/icde/LiDCH22}.
Another promising avenue is integrating the concept of neighborhoods among clients, 
 as certain algorithms (e.g., PBFL~\cite{DBLP:conf/acsos/DominiAFVE24}) leverage distributed 
 localized networks for cluster formation.

\section*{Acknowledgment}
Gianluca Aguzzi was supported by the Italian PRIN project ``CommonWears'' (2020 HCWWLP) 
 and the EU/MUR FSE PON-R \&I 2014-2020.

\bibliographystyle{IEEEtran}
\bibliography{IEEEexample}

\end{document}